\documentclass{article}

\usepackage{arxiv}

\usepackage[utf8]{inputenc}     
\usepackage[T1]{fontenc}        
\usepackage{hyperref}           
\usepackage{url}                
\usepackage{booktabs}           
\usepackage{amsfonts}           
\usepackage{amsmath}            
\usepackage{nicefrac}           
\usepackage{microtype}          
\usepackage{graphicx}           
\usepackage{xcolor}             
\usepackage{listings}           
\usepackage{array}              
\usepackage[utf8]{inputenc}
\hypersetup{
    colorlinks=true,
    linkcolor=blue,
    citecolor=blue,
    urlcolor=blue
}

\title{600K-KS-OCR: A Large-Scale Synthetic Dataset for \\Optical Character Recognition in Kashmiri Script}

\author{
    Haq Nawaz Malik \\
    Independent Researcher \\
     \texttt {orcid.org/0009-0003-1994-7640} \\
    \texttt{huggingface.co/Omarrran}\\
     \texttt{x.com/HAQ\_NAWAZ\_MALIK}   \\ 
}

\begin{document}

\maketitle

\begin{abstract}
This technical report presents the \textbf{600K-KS-OCR Dataset}, a large-scale synthetic corpus comprising approximately 602,000 word-level segmented images designed for training and evaluating optical character recognition systems targeting Kashmiri script. The dataset addresses a critical resource gap for Kashmiri, an endangered Dardic language utilizing a modified Perso-Arabic writing system spoken by approximately seven million people. Each image is rendered at $256 \times 64$ pixels with corresponding ground-truth transcriptions provided in multiple formats compatible with CRNN, TrOCR, and general-purpose machine learning pipelines. The generation methodology incorporates three traditional Kashmiri typefaces, comprehensive data augmentation simulating real-world document degradation, and diverse background textures to enhance model robustness. The dataset is distributed across ten partitioned archives totaling approximately 10.6 GB and is released under the CC-BY-4.0 license to facilitate research in low-resource language optical character recognition.
\end{abstract}

\keywords{Optical Character Recognition \and Kashmiri Script \and Low-Resource Languages \and Synthetic Dataset \and Deep Learning \and Perso-Arabic Script}

\section{Introduction}
\label{sec:introduction}

Optical Character Recognition (OCR) systems have achieved remarkable accuracy on high-resource languages with abundant annotated training data. However, performance degrades substantially for low-resource languages where labeled datasets are scarce or nonexistent. This disparity is particularly acute for languages employing complex calligraphic scripts with limited existing computational infrastructure.

Kashmiri , the principal language of the Kashmir Valley, exemplifies these challenges. As a Dardic language of the Indo-Aryan family, Kashmiri employs a modified Perso-Arabic script containing additional characters to represent phonemes absent in Arabic and Persian \cite{koul2006kashmiri}. Despite a speaker population of approximately seven million, Kashmiri remains severely underrepresented in natural language processing and computer vision research.

The development of effective OCR systems for Kashmiri script is essential for multiple applications: digitization of historical manuscripts and administrative records, preservation of literary heritage, and enablement of text-based accessibility technologies. However, the absence of large-scale annotated datasets has impeded progress in this area.

This technical report presents the 600K-KS-OCR Dataset, a synthetic corpus of approximately 602,000 word-level segmented images with corresponding ground-truth transcriptions. The dataset was designed with the following objectives:

\begin{enumerate}
    \item \textbf{Scale}: Provide sufficient training data for deep learning architectures that require large sample sizes to achieve robust generalization.
    
    \item \textbf{Authenticity}: Employ traditional Kashmiri typefaces that accurately represent the script's calligraphic characteristics.
    
    \item \textbf{Robustness}: Incorporate comprehensive augmentation to simulate real-world document conditions including noise, blur, geometric distortion, and paper degradation.
    
    \item \textbf{Accessibility}: Distribute the dataset in multiple formats compatible with prevalent OCR training frameworks.
\end{enumerate}

The remainder of this report is organized as follows. Section~\ref{sec:background} provides background on Kashmiri script characteristics and related work. Section~\ref{sec:dataset} details dataset specifications and architecture. Section~\ref{sec:methodology} describes the generation methodology. Section~\ref{sec:formats} specifies data format conventions. Section~\ref{sec:applications} discusses intended applications. Section~\ref{sec:conclusion} concludes with availability information.

\section{Background and Related Work}
\label{sec:background}

\subsection{Kashmiri Script Characteristics}

The Kashmiri writing system presents several challenges for computational recognition:

\paragraph{Perso-Arabic Foundation.} Kashmiri script is derived from the Perso-Arabic alphabet but extends it with additional characters to represent sounds unique to the language. These include modified forms of existing letters and entirely new graphemes with distinctive diacritical marks.

\paragraph{Right-to-Left Directionality.} Text flows from right to left, requiring appropriate handling in image rendering and sequence modeling architectures.

\paragraph{Contextual Letter Forms.} Characters assume different shapes depending on their position within a word (initial, medial, final, or isolated), a characteristic shared with Arabic and Persian scripts.

\paragraph{Complex Diacritics.} Kashmiri employs an extensive system of diacritical marks (vowel signs, shadda, sukun) that modify pronunciation and meaning. These marks are positioned above or below base characters and must be accurately rendered and recognized.

\paragraph{Ligatures and Joining.} Adjacent characters connect in calligraphic traditions, forming ligatures that alter the visual appearance of letter sequences.

\subsection{Related Work}

OCR research for Perso-Arabic scripts has primarily focused on Arabic, Persian, and Urdu. Notable datasets include the IFN/ENIT database for Arabic handwriting \cite{pechwitz2002ifn}, the UPTI dataset for Urdu printed text \cite{sabbour2013database}, and the KHATT database for Arabic handwritten text \cite{mahmoud2014khatt}.

For Kashmiri specifically, computational resources remain extremely limited. Prior work has addressed isolated character recognition \cite{ahmad2017kashmiri} but large-scale word-segmented datasets suitable for end-to-end OCR training have been unavailable.

Synthetic data generation has proven effective for augmenting OCR training corpora. SynthText \cite{gupta2016synthetic} demonstrated the utility of rendered text images for scene text recognition. The MJSynth dataset \cite{jaderberg2014synthetic} provided millions of synthetic word images that enabled breakthrough performance on benchmark datasets. This work extends the synthetic data paradigm to Kashmiri script.

\section{Dataset Specifications}
\label{sec:dataset}

\subsection{Overview}

Table~\ref{tab:overview} summarizes the primary characteristics of the 600K-KS-OCR Dataset.

\begin{table}[h]
\caption{Dataset Overview}
\centering
\begin{tabular}{ll}
\toprule
\textbf{Property} & \textbf{Specification} \\
\midrule
Total Samples & $\sim$602,000 words \\
Archive Distribution & 10 ZIP files \\
Total Size & $\sim$10.6 GB \\
Image Dimensions & $256 \times 64$ pixels \\
Image Format & PNG (lossless) \\
Text Direction & Right-to-Left (RTL) \\
Script System & Kashmiri (Perso-Arabic) \\
Output Formats & CRNN, TrOCR, CSV, JSONL \\
License & CC-BY-4.0 \\
\bottomrule
\end{tabular}
\label{tab:overview}
\end{table}

\subsection{Partition Structure}

The dataset is distributed across ten partitioned archives to facilitate manageable downloads and modular usage. Table~\ref{tab:partitions} details the partition specifications.

\begin{table}[h]
\caption{Dataset Partitions}
\centering
\begin{tabular}{lrr}
\toprule
\textbf{Partition} & \textbf{Samples} & \textbf{Size} \\
\midrule
P1\_OCR\_dataset & 50,000 & $\sim$904 MB \\
P2\_OCR\_dataset & 53,815 & $\sim$953 MB \\
P3\_OCR\_dataset & 68,741 & $\sim$1.2 GB \\
P4\_OCR\_dataset & 69,886 & $\sim$1.2 GB \\
P5\_OCR\_dataset & 69,637 & $\sim$1.2 GB \\
P6\_OCR\_dataset & 69,506 & $\sim$1.2 GB \\
P7\_OCR\_dataset & 58,228 & $\sim$1.1 GB \\
P8\_OCR\_dataset & 35,720 & $\sim$620 MB \\
P9\_OCR\_dataset & 86,635 & $\sim$1.5 GB \\
P10\_OCR\_dataset & 41,401 & $\sim$732 MB \\
\midrule
\textbf{Total} & \textbf{$\sim$602,000} & \textbf{$\sim$10.6 GB} \\
\bottomrule
\end{tabular}
\label{tab:partitions}
\end{table}

\subsection{Archive Contents}

Each partition archive maintains a consistent internal structure:

\begin{lstlisting}[language=bash,caption={Archive Directory Structure}]
P*_OCR_dataset_*/
+-- images/           # Word-segmented PNG images
+-- data.csv          # Tabular format (filename, text)
+-- data.jsonl        # JSON Lines for TrOCR pipelines
+-- labels.txt        # Tab-separated CRNN format
+-- metadata.json     # Generation configuration
\end{lstlisting}

This organization enables researchers to select their preferred label format while maintaining consistent image references across formats.

\section{Generation Methodology}
\label{sec:methodology}

\subsection{Rendering Configuration}

All images were rendered with standardized parameters optimized for Convolutional Recurrent Neural Network (CRNN) and Transformer-based OCR architectures:

\begin{table}[h]
\caption{Rendering Parameters}
\centering
\begin{tabular}{ll}
\toprule
\textbf{Parameter} & \textbf{Value} \\
\midrule
Image Dimensions & $256 \times 64$ pixels \\
Background Color & \#FFFFFF (white) \\
Text Color & \#000000 (black) \\
Text Direction & Right-to-Left \\
Background Mode & Mixed (varied textures) \\
Color Space & RGB \\
\bottomrule
\end{tabular}
\label{tab:rendering}
\end{table}

The $256 \times 64$ pixel dimension was selected to balance resolution quality against computational efficiency, representing a common configuration for word-level OCR systems \cite{shi2017crnn}.

\subsection{Typography}

Text rendering employed three typefaces representing traditional Kashmiri and Nastaliq calligraphic styles:

\paragraph{Afan Koshur Naksh.} A native Kashmiri Naskh-style typeface providing clear, readable letterforms suitable for printed materials. This font emphasizes legibility while maintaining authentic Kashmiri character shapes.

\paragraph{Nastaleeq.} A classical calligraphic typeface exhibiting the diagonal baseline and flowing connections characteristic of formal literary and religious Kashmiri texts.

\paragraph{Nakash (Narqalam).} A typeface incorporating traditional handwritten qualities that simulate manuscript and informal document conditions, introducing natural variation absent from purely typeset fonts.

The use of multiple typefaces enhances model generalization across the stylistic diversity encountered in real-world Kashmiri documents.

\subsection{Data Augmentation Pipeline}

To promote robustness against real-world image degradation, a comprehensive augmentation pipeline was applied to 60\% of generated samples. The remaining 40\% were retained as clean baseline images. Augmentation categories include:

\paragraph{Geometric Transformations.} Rotation (±5°), perspective distortion, skew, and tilt variations simulate scanning misalignment and document positioning artifacts.

\paragraph{Blur Effects.} Gaussian blur and motion blur simulate focus variations and movement during image capture.

\paragraph{Noise Injection.} Gaussian noise and salt-and-pepper patterns characteristic of low-quality digitization equipment and sensor noise.

\paragraph{Photometric Variations.} Brightness and contrast adjustment, JPEG compression artifacts, and resolution degradation simulate diverse capture conditions.

\paragraph{Document-Specific Effects.} Paper texture overlay, shadow and lighting gradients, and ink bleed phenomena commonly observed in aged or degraded source materials.

\subsection{Background Synthesis}

The mixed background mode incorporates diverse textures to improve generalization beyond pristine white backgrounds:

\begin{itemize}
    \item \textbf{Clean backgrounds}: Pure white for optimal baseline conditions
    \item \textbf{Aged document styles}: Aged paper, antiquarian book pages, parchment
    \item \textbf{Document variants}: Notebook paper, printed book pages, newspaper textures
    \item \textbf{Paper tones}: Ivory, cream, recycled paper appearances
    \item \textbf{Distressed effects}: Coffee stains, water damage, weathered surfaces
    \item \textbf{Custom textures}: Fourteen additional background variations
\end{itemize}

This background diversity addresses the domain shift between synthetic training data and heterogeneous real-world documents.

\subsection{Generation Infrastructure}

Dataset generation employed GPU-accelerated parallel processing across four computational cores, achieving 3-10$\times$ performance improvements over sequential generation. GPU-based augmentation enabled efficient application of complex transformations at scale.

\section{Data Format Specifications}
\label{sec:formats}

The dataset provides labels in four formats to accommodate diverse training frameworks and workflows.

\subsection{CRNN Format (labels.txt)}

Tab-separated format following the standard convention for Connectionist Temporal Classification (CTC) based sequence recognition:

\begin{lstlisting}[caption={CRNN Label Format}]
image_001.png	Kashmiri Text
image_002.png	Kashmiri Text
image_003.png	Kashmiri Text
\end{lstlisting}

\begin{table}[h]
\centering
\begin{tabular}{|c|c}
\hline
\textbf{Image} & \\
\hline
\includegraphics[width=0.4\textwidth]{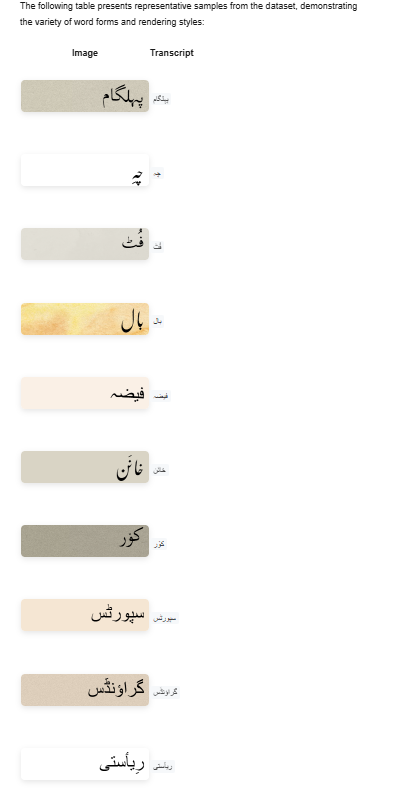} & \\

\hline
\end{tabular}
\caption{Kashmiri OCR Dataset Samples}
\label{tab:kashmiri_ocr}
\end{table}

\subsection{TrOCR Format (data.jsonl)}

JSON Lines format enables direct integration with Hugging Face Transformers pipelines for TrOCR and similar encoder-decoder architectures:

\begin{lstlisting}[caption={TrOCR Label Format}]
{"file_name": "image_001.png", "text": "Kashmiri Text"}
{"file_name": "image_002.png", "text": "Kashmiri Text"}
{"file_name": "image_003.png", "text": "Kashmiri Text"}
\end{lstlisting}

\subsection{Tabular Format (data.csv)}

Standard CSV format provides compatibility with general-purpose data processing tools:

\begin{lstlisting}[caption={CSV Label Format}]
filename,text
image_001.png,Kashmiri Text
image_002.png,Kashmiri Text
image_003.png,Kashmiri Text
\end{lstlisting}

\subsection{Metadata Schema}

Each archive includes comprehensive metadata documenting generation parameters:

\begin{lstlisting}[caption={Metadata JSON Schema}]
{
  "generated_at": "2025-12-26T16:24:34.803Z",
  "config": {
    "image_size": "256x64",
    "augmentation_enabled": true,
    "augmentation_percentage": 60,
    "fonts_used": [
      "Afan_Koshur_Naksh",
      "Nastaleeq",
      "Nakash"
    ],
    "output_formats": ["crnn", "trocr", "csv", "jsonl"]
  },
  "samples": 50000,
  "clean_samples": 20000,
  "augmented_samples": 30000
}
\end{lstlisting}

\section{Applications and Impact}
\label{sec:applications}

\subsection{Primary Applications}

The 600K-KS-OCR Dataset supports multiple research and development applications:

\paragraph{OCR Model Training.} The dataset provides sufficient scale for training deep learning architectures including CRNN \cite{shi2017crnn}, TrOCR \cite{li2023trocr}, and attention-based sequence-to-sequence models from scratch or via fine-tuning.

\paragraph{Benchmarking.} Standardized partitions and format specifications enable consistent evaluation of OCR systems on Kashmiri script, facilitating reproducible research comparisons.

\paragraph{Transfer Learning.} Models pretrained on this dataset may transfer to related Perso-Arabic scripts or serve as initialization for domain-specific Kashmiri OCR applications.

\paragraph{Document Digitization.} Practical deployment of trained models enables digitization of historical manuscripts, newspapers, administrative records, and literary archives currently inaccessible in machine-readable form.

\subsection{Broader Impact}

Development of robust Kashmiri OCR capabilities contributes to language preservation efforts for this endangered language. Digitization of textual heritage enables:

\begin{itemize}
    \item Archival preservation of deteriorating physical documents
    \item Full-text search across digitized collections
    \item Accessibility technologies for visually impaired readers
    \item Corpus construction for natural language processing research
\end{itemize}

\subsection{Usage Example}

The dataset integrates directly with the Hugging Face ecosystem:

\begin{lstlisting}[language=Python,caption={Loading via Hugging Face Datasets}]
from datasets import load_dataset

dataset = load_dataset(
    "Omarrran/600k_KS_OCR_Word_Segmented_Dataset"
)
\end{lstlisting}

For manual loading:

\begin{lstlisting}[language=Python,caption={Manual Data Loading}]
import pandas as pd
from PIL import Image

labels_df = pd.read_csv("data.csv")
img = Image.open("images/image_001.png")
text = labels_df.loc[
    labels_df['filename'] == 'image_001.png', 
    'text'
].values[0]
\end{lstlisting}

\section{Limitations}
\label{sec:limitations}

Several limitations of this dataset should be acknowledged:

\paragraph{Synthetic Nature.} As a synthetically generated corpus, the dataset may not fully capture the variability of authentic handwritten or degraded historical documents. Performance on real-world data should be validated through domain adaptation or fine-tuning on genuine document samples when available.

\paragraph{Word-Level Segmentation.} The dataset provides pre-segmented word images. Systems requiring line-level or page-level recognition will need additional datasets or synthetic generation for those granularities.

\paragraph{Font Coverage.} While three typefaces provide stylistic diversity, additional fonts particularly those representing regional calligraphic variations would further enhance generalization.

\paragraph{Vocabulary Distribution.} The vocabulary distribution reflects the source text corpus used for rendering. Specialized domains (technical, medical, legal) may require supplementary data.

\section{Conclusion}
\label{sec:conclusion}

This technical report has presented the 600K-KS-OCR Dataset, a large-scale synthetic corpus for Kashmiri optical character recognition comprising approximately 602,000 word-segmented images with ground-truth transcriptions. The dataset addresses a critical resource gap for Kashmiri, an underrepresented language in computational vision research.

Key characteristics of the dataset include standardized $256 \times 64$ pixel images rendered with traditional Kashmiri typefaces, comprehensive augmentation simulating real-world document degradation, diverse background textures, and distribution in multiple formats compatible with prevalent OCR training frameworks.

The dataset is released under the CC-BY-4.0 license to facilitate research in low-resource language OCR. Future work may extend this resource through additional fonts, expanded vocabulary coverage, and line-level or page-level synthetic generation.

\section*{Data Availability}

The 600K-KS-OCR Dataset is available through the Hugging Face Datasets Hub:

\begin{center}
\url{https://huggingface.co/datasets/Omarrran/600k_KS_OCR_Word_Segmented_Dataset}
\end{center}

\noindent The dataset is released under the CC-BY-4.0 license permitting use with appropriate attribution.

\section*{Acknowledgments}

The author acknowledges the Kashmiri language community for linguistic guidance and the developers of open-source typefaces used in this dataset.

\bibliographystyle{unsrt}

\begin{thebibliography}{10}

\bibitem{koul2006kashmiri}
Omkar~N. Koul and Kashi Wali.
\newblock \emph{Modern Kashmiri Grammar}.
\newblock Dunwoody Press, 2006.

\bibitem{pechwitz2002ifn}
Mario Pechwitz, S.~Snoussi Maddouri, Volker M{\"a}rgner, Noureddine Ellouze, and Haikal Amiri.
\newblock {IFN/ENIT} - Database of Handwritten {A}rabic Words.
\newblock In \emph{Proc. CIFED}, volume~2, pages 127--136, 2002.

\bibitem{sabbour2013database}
Nazly Sabbour and Faisal Shafait.
\newblock A Database for {U}rdu Printed Text Recognition.
\newblock In \emph{Document Analysis and Recognition (ICDAR)}, pages 1339--1343. IEEE, 2013.

\bibitem{mahmoud2014khatt}
Sabri~A. Mahmoud, Irfan Ahmad, Wasfi~G. Al-Khatib, Mohammad Alshayeb, Mohammad~Tanvir Parvez, Volker M{\"a}rgner, and Gernot~A. Fink.
\newblock {KHATT}: An Open {A}rabic Offline Handwritten Text Database.
\newblock \emph{Pattern Recognition}, 47(3):1096--1112, 2014.

\bibitem{ahmad2017kashmiri}
Sheeraz Ahmad and Syed Arsalan.
\newblock Recognition of Isolated Kashmiri Characters.
\newblock In \emph{International Conference on Computing and Communication Technologies}, pages 1--5, 2017.

\bibitem{gupta2016synthetic}
Ankush Gupta, Andrea Vedaldi, and Andrew Zisserman.
\newblock Synthetic Data for Text Localisation in Natural Images.
\newblock In \emph{Proceedings of the IEEE Conference on Computer Vision and Pattern Recognition}, pages 2315--2324, 2016.

\bibitem{jaderberg2014synthetic}
Max Jaderberg, Karen Simonyan, Andrea Vedaldi, and Andrew Zisserman.
\newblock Synthetic Data and Artificial Neural Networks for Natural Scene Text Recognition.
\newblock In \emph{NIPS Deep Learning Workshop}, 2014.

\bibitem{shi2017crnn}
Baoguang Shi, Xiang Bai, and Cong Yao.
\newblock An End-to-End Trainable Neural Network for Image-based Sequence Recognition and Its Application to Scene Text Recognition.
\newblock \emph{IEEE Transactions on Pattern Analysis and Machine Intelligence}, 39(11):2298--2304, 2017.

\bibitem{li2023trocr}
Minghao Li, Tengchao Lv, Jingye Chen, Lei Cui, Yijuan Lu, Dinei Florencio, Cha Zhang, Zhoujun Li, and Furu Wei.
\newblock {TrOCR}: Transformer-based Optical Character Recognition with Pre-trained Models.
\newblock In \emph{Proceedings of the AAAI Conference on Artificial Intelligence}, volume~37, pages 13094--13102, 2023.

\end{thebibliography}

\end{document}